\documentstyle{article}

\title{AN EXPERIMENT ON LEARNING APPROPRIATE SELECTIONAL
RESTRICTIONS FROM A PARSED CORPUS\footnote{this is a revised version
of the original paper published in the Proceedings of the 15th
International Conference on Computational Linguistics, August 1994,
Kyoto, Japan.}  \\}
\author{Francesc Ribas Framis\thanks{The reported research has been
supported by a grant conceded by the Generalitat de Catalunya,
91-DOGC-1491.  Much of the work reported here was carried out during a
visit at the Computer Laboratory, University of Cambridge. I am
grateful to Ted Briscoe and Horacio Rodriguez by their valuable
comments.}\\ \\ Departament de Llenguatges i Sistemes Inform\`{a}tics,
\\ Universitat Polit\`{e}cnica de Catalunya\\ Pau Gargallo 5, 08082
Barcelona, SPAIN. \\ e-mail: ribas@lsi.upc.es}
\date{Archive number: cmp-lg/9409004}

\begin{document}

\bibliographystyle{alpha}

\maketitle

\begin{abstract}

We present a methodology to extract Selectional Restrictions at a
variable level of abstraction from phrasally analyzed corpora. The
method relays in the use of a wide-coverage noun taxonomy and a
statistical measure of the co-occurrence of linguistic items. Some
experimental results about the performance of the method are
provided.

\end{abstract}

\section{\bf \sc Introduction}
\label{introduction}
These last years there has been a common agreement in the natural
language processing research community on the importance of having an
extensive coverage of the surface lexical semantics of the domain to
work with, (specially, typical contexts of use). This knowledge may be
expressed at different levels of abstraction depending on the
phenomena involved: selectional restrictions (SRs), lexical
preferences, col-locations, etc. We are specially interested on SRs,
which can be expressed as semantic type constraints that a word sense
imposes on the words with which it combines in the process of semantic
interpretation. SRs must include information on the syntactic position
of the words that are being restricted semantically.  For instance,
one of the senses of the verb {\it drink} restricts its {\it subject}
to be an {\it animal} and its object to be a {\it liquid}.

SRs may help a parser to prefer some parses among several grammatical
ones
\cite{whittemore-90}.  Furthermore, SRs may help the parser when deciding the
semantic role played by a syntactic complement. Lexicography is also
interested in the acquisition of SRs. On the one hand, SRs are an
interesting information to be included in dictionaries ({\it defining
in context} approach). On the other hand, as \cite{church-90} remark,
the effort involved in analyzing and classifying all the linguistic
material provided by concordances of use of a word can be extremely
labor-intensive. If it was possible to represent roughly the SRs of
the word being studied, it could be possible to classify roughly the
concordances automatically in the different word uses before the
lexicographer analysis.

The possible sources of SRs are: introspection by lexicographers,
machine-readable dictionaries, and on-line corpora.  The main
advantage of the latter is that they provide experimental evidence of
words uses. Recently, several approaches on acquiring different kinds
of lexical information from corpora have been developed
\label{corpus-methods}
\cite{basili-92a,church-91,church-90,resnik-92}. This paper is
interested in exploring the amenability of using a method for
extracting SRs from textual data, in the line of these works.  The aim
of the proposed technique is to learn the SRs that a word is imposing,
from the analysis of the examples of use of that word contained in the
corpus. An illustration of such a learning is shown in Figure
\ref{figure-examples}, where the system, departing from the three
examples of use, and knowing that {\it prosecutor}, {\it buyer} and
{\it lawmaker} are nouns belonging to the semantic class
\mbox{$<\!person, individual\!>$}, and that {\it indictment}, {\it
assurance} and {\it legislation} are members of
\mbox{$<\!legal\_instrument\!>$}, should induce that the verb {\it
seek} imposes SRs that constraint the subject to be a member
of the semantic type \mbox{$<\!person, individual\!>$}, and the object
to be a kind of \mbox{$<\!legal\_instrument\!>$}.  Concluding, the
system should extract for each word (with complements) having enough
number occurrences of use in the corpus and for each of its syntactic
complements, a list of the alternative SRs that this word is imposing.

\begin{figure}
\begin{itemize}

\item {\bf Three examples of use}

{\it prosecutors} may soon {\it seek} an {\it indictment} on
racketeering and securities fraud charges.

In the recent past, bond {\it buyers} didn't {\it seek} such {\it
assurance}.

Some {\it lawmakers} may {\it seek} {\it legislation} to limit overly
restrictive insurance policies.

\item {\bf The extracted SRs}

 {\it (seek, subject, \mbox{$<\!person, individual\!>$})}

{\it (seek, object, \mbox{$<\!legal\_instrument\!>$})}

\end{itemize}
\caption{Example of the acquisition of SRs for
the verb {\it seek} from three examples of use}
\label{figure-examples}
\end{figure}

In order to detect the SRs that a word imposes in its context by means
of statistical techniques two distinct approaches have been proposed:
{\it word-based} \cite{church-91}, and {\it class-based}
\cite{basili-92a,resnik-92}. Word-based approach infers SRs as the
collection of words that co-occur significantly in the syntactic
context of the studied word.  The class-based techniques gather the
different nouns by means of semantic classes. The advantages of the
latter are clear. On the one hand, statistically meaningful data can
be gathered from (relatively) small corpora,and not only for the most
frequent words.  On the other hand, SRs are generalized to new
examples not present in the training set. Finally, the acquired SRs
are more independent of the lexical choices made in the training
corpus.

We have developed and implemented a method for automatically
extracting class-based SRs from on-line corpora.  In section
\ref{method-acquisition} we describe it while discussing other
approaches. In section
\ref{experimental-results} we analyze some data about the
performance of an experiment run in a Unix machine, on a corpus of
800,000 words.  Finally, in section \ref{conclusions} we discuss the
performance achieved, and suggest further refinements of the technique
in order to solve some remaining problems.

\section{\bf \sc The Method of Acquisition}
\label{method-acquisition}

SRs have been used to express semantic constraints holding in
different syntactic and functional configurations. However, in this
paper we focus only in selectional restrictions holding between verbs
and their complements. The method can be easily exported to other
configurations. We won't distinguish the SRs imposed by verbs on
arguments and adjuncts. We believe that few adjuncts are going to
provide enough evidence in the corpus for creating SRs. In the
following paragraphs we describe the functional specification of the
system.

\begin{description}

\item[Training set\\]

The input to the learning process is a list of {\it co-occurrence
triples} codifying the co-occurrence of verbs and complement heads in
the corpus: {\it (verb, syntactic relationship, noun)}.  {\it Verb}
and {\it noun} are the lemmas of the inflected forms appearing in
text. {\it Syntactic relationship} codes the kind of complement: {\it
0} subject, {\it 1} object , or {\it preposition} in case it is a
PP. A method to draw the co-occurrence triples from corpus is proposed
in subsection \ref{extracting-triples}.

\item[Output\\]

The result of the learning process is a set of syntactic SRs, {\it
(verb, syntactic relationship, semantic class)}.  Semantic classes are
represented extensionally as sets of nouns.  SRs are only acquired if
there are enough cases in the corpus as to gather statistical
evidence.  As long as distinct uses of the same verb can have
different SRs, we permit to extract more than one class for the same
syntactic position. Nevertheless, they must be mutually disjoint,
i.e. not related by hyperonymy.

\item[Previous knowledge used\\]

In the process of learning SRs, the system needs to know how words are
clustered in semantic classes, and how semantic classes are
hierarchically organized. Ambiguous words must be represented as
having different hyperonym classes.  In subsection
\ref{semantic-knowledge} we defend the use of a broad-coverage
taxonomy.

\item[Learning process\\]

The computational process is divided in three stages: (1) Guessing the
possible semantic classes, i.e.  creation of the space of candidates.
In principle, all the hyperonyms (at all levels) of the nouns
appearing in the training set are candidates.  (2) Evaluation of the
appropriateness of the candidates. In order to compare the different
candidates, an statistical measure summarizing the relevance of the
occurrence of each of the candidate classes is used.  (3) Selection of
the most appropriate subset of the candidate space to convey the SRs,
taking into account that the final classes must be mutually disjoint.
While in subsection \ref{association-score} an statistical measure to
fulfill stage 2 is presented, stages 1 and 3 are discussed in
\ref{best-classes} thoroughly.

\end {description}

\subsection{Extracting Co-occurrence Triples}
\label{extracting-triples}

In any process of learning from examples the accuracy of the training
set is the base for the system to make correct predictions. In our
case, where the semantic classes are hypothesized not univoquely from
the examples, accuracy becomes fundamental.

Different approaches to ob\-tain le\-xi\-cal co-oc\-cur\-ren\-ces have
been proposed in the literature \cite{basili-92a,church-91,church-90}.
These approaches seem inappropriate for tackling our needs, either
because they detect only local
co-occurrences\cite{church-91,church-90}, or because they extract many
spurious co-occurrence triples \cite{basili-92a,church-90}.  On the
one hand, our system intends to learn SRs on any kind of verb's
complements. On the other hand, the fact that these approaches extract
co-occurrences without reliability on being verb-complements violates
accuracy requirements.

However, if the co-occurrences were extracted from a corpus annotated
with structural syntactic information (i.e., part of speech and
``skeletal'' trees), the results would have considerably higher
degrees of accuracy and representativity. In this way, it would be
easy to detect all the relationships between verb and complements, and
few non-related co-occurrences would be extracted. The most serious
objection to this approach is that the task of producing syntactic
analyzed corpora is very expensive.  Nevertheless, lately there has
been a growing interest to produce skeletally analyzed
corpora\footnote{For instance, Penn Treebank Corpus, which is being
collected and analyzed by the University of Pennsylvania (see
\cite{marcus-93}). The material is available on request from the
Linguistic Data Consortium, (email) ldc@unagi.cis.upenn.edu}

A parser, with some simple heuristics, would be enough to meet the
requirements of representativeness and accuracy introduced above. On
the other hand, it could be useful to represent the co-occurrence
triples as holding between lemmas, in order to gather as much evidence
as possible. A simple morphological analyzer that could get the lemma
for a big percentage of the words appearing in the corpus would
suffice.

\subsection{Semantic Knowledge Used}
\label{semantic-knowledge}

Of the two class-based approaches presented in section
\ref{introduction}, \cite{resnik-92}'s technique uses a wide-coverage
semantic taxonomy , whereas
\cite{basili-92a} consists in hand-tagging with a fixed set of semantic
labels . The advantages and drawbacks of both approaches are
diverse. On the one hand, in \cite{basili-92a} approach, semantic
classes relevant to the domain are chosen, and consequently, the
adjustment of the classes to the corpus is quite nice. Nevertheless,
\cite{resnik-92}'s system is
less constrained and is able to induce a most appropriate level for
the SRs. On the other hand, while \cite{basili-92a} implies
hand-coding all the relevant words with semantic tags,
\cite{resnik-92} needs a broad semantic taxonomy. However, there is
already an available taxonomy, WordNet\footnote{WordNet is a lexical
database developed with psycholinguistic aims.  It represents lexical
semantics information about nouns, verbs, adjectives and adverbs such
as hyperonyms, meronyms, ... It presently contains information on
about 83,000 lemmas. See
\cite{miller-90}}. We take \cite{resnik-92} approach because
of the better results obtained, and the lower cost involved.

\subsection{Class appropriateness: the Association Score}
\label{association-score}

When trying to choose a measure of the appropriateness of a semantic
class, we have to consider the features of the problem: (1) robustness
in front of noise, and (2) conservatism in order to be able to
generalize only from positive examples, without having the tendency to
over-generalize.

Several statistical measures that accomplish these requirements have
been proposed in the literature \cite{basili-92a,church-91,resnik-92}.
We adopt \cite{resnik-92}'s approach, which quantifies the statistical
association between verbs and classes of nouns from their
co-occurrence. However we adapt it taking into account the syntactic
position of the relationship. Let

\[ {\cal V} = \{ v_{1}, \ldots, v_{l} \}, \: {\cal N} = \{ n_{1},
\ldots, n_{m} \}, \]
\[{\cal S} = \{ 0, 1, to, at, \ldots \}, \: and \: {\cal C} = \{
c|c \subseteq {\cal N} \} \]

be the sets of all verbs, nouns, syntactic positions, and possible
noun classes, respectively.  Given \(v \in {\cal V}
\), \(s \in {\cal S} \) and \(c \in {\cal C} \), {\it Association
Score}, $Assoc$, between $v$ and $c$ in a syntactic position $s$ is
defined to be

\[
\label{eq-assoc-score}
Assoc(v,s,c) \equiv P(c|v,s) I(v;c|s) \]
\[ = P(c|v,s) \log_{2}
\frac{P(v,c|s)}{P(v|s)P(c|s)}
\]

Where conditional probabilities are estimated by counting the number
of observations of the joint event and dividing by the frequency of
the given event, e.g.
\[ P(v,c|s) \approx \frac{\sum_{n \in c} count(v,s,n)}
        {\sum_{v' \in {\cal V}} \sum_{n' \in {\cal N}} count(v',s,n')}
\]

The two terms of Assoc try to capture different properties of the SR
expressed by the candidate class. Mutual information, $I(v;c|s)$,
measures the strength of the statistical association between the given
verb $v$ and the candidate class $c$ in the given syntactic position
$s$.  If there is a real relationship, then hopefully $I(v,c|s)
\gg 0$.  On the other hand, the conditional probability, $P(c|v,s)$,
favors those classes that have more occurrences of nouns.

\subsection{Selecting the best classes}
\label{best-classes}

The existence of noise in the training set introduces classes in the
candidate space that can't be considered as expressing SRs.  A common
technique used for ignoring as far as possible this noise is to
consider only those events that have a higher number of occurrences
than a certain {\it threshold}. However, some erroneous classes may
persist because they exceed the threshold. However, if candidate
classes were ordered by the significance of their Assoc with the verb,
it is likely that less appropriate classes (introduced by noise) would
be ranked in the last positions of the candidate list.

The algorithm to learn SRs is based in a search through all the
classes with more instances in the training set than the given
threshold. In different iterations over these candidate classes, two
operations are performed: first, the class, $c$, having the best
$Assoc$ (best class), is extracted for the final result; and second,
the remaining candidate classes are filtered from classes being
hyper/hyponyms to the best class.  This last step is made because the
definitive classes must be mutually disjoint. The iterations are
repeated until the candidate space has been run out.

\cite{resnik-92} performed a similar learning process, but while he was only
looking for the preferred class of object nouns, we are interested in
all the possible classes (SRs). He performed a best-first search on
the candidate space. However, if the function to maximize doesn't have
a monotone behavior (as it is the case of Assoc) the best-first search
doesn't guarantee global optimals, but only local ones.  This fact
made us to decide for a global search, specially because the candidate
space is not so big.

\section{\bf \sc Experimental Results}
\label{experimental-results}

In order to experiment the methodology presented, we implemented a
system in a Unix machine.  The corpus used for extracting
co-occurrence triples is a fragment of parsed material from the Penn
Treebank Corpus (about 880,000 words and 35,000 sentences), consisting
of articles of the Wall Street Journal, that has been tagged and
parsed.  We used Wordnet as the verb and noun lexicons for the
lemmatizer, and also as the semantic taxonomy for clustering nouns in
semantic classes. In this section we evaluate the performance of the
methodology implemented: (1) looking at the performance of the
techniques used for extracting triples, (2) considering the coverage
of the WordNet taxonomy regarding the noun senses appearing in
Treebank, and (3) analyzing the performance of the learning process.

\begin{table}
\centering

\begin{tabular}{||l|c|r|r|r|l||} \hline
{\it Acquired SR } & {\it Type} & {\it Assoc} & \#$n$ & \#$s$ & {\it
Examples of nouns in Treebank}\\ \hline $<\!  cognition \!>$ & Senses
& -0.04 & 5 & 1 & concern, leadership, provision, science \\ $<\!
activity \!>$ & Senses & -0.01 & 6 & 1 & administration, leadership,
provision \\ $<\!  status \!>$ & Senses & 0.087 & 5 & 0 & government,
leadership
\\ $<\!
social\_control \!>$ & Senses & 0.11 & 6 & 0 & administration,
government
\\ $<\!
administrative\_district \!>$ & Senses & 0.14 & 36 & 0 & proper\_name
\\ $<\!  city \!>$ & Senses & 0.15 & 36 & 0 & proper\_name \\ $<\!
radical \!>$ & Senses & 0.16 & 5 & 0 & group \\ $<\!  person,
individual \!>$ & Ok & 0.23 & 61 & 38 & advocate, buyer, carrier,
client, ... \\ $<\!  legal\_action \!>$ & Ok & 0.28 & 7 & 6 & suit \\
$<\!  group \!>$ & $\Uparrow$Abs. & 0.35 & 64 & 46 & administration,
agency, bank, ... \\ $<\!  suit \!>$ & Senses & 0.40 & 7 & 0 & suit \\
$<\!  suit\_of\_clothes \!>$ & Senses & 0.41 & 7 & 0 & suit \\ $<\!
suit, suing \!>$ & Senses & 0.41 & 7 & 0 & suit \\ \hline
\end{tabular}

\caption{SRs acquired for the subject of {\it seek} }
\label{seek-1-table}

\end{table}

The total number of co-occurrence triples extracted amounts to
190,766. Many of these triples (68,800, 36.1\%) were discarded before
the lemmatizing process because the surface NP head wasn't a noun. The
remaining 121,966 triples were processed through the lemmatizer.
113,583 (93.1\%) could be correctly mapped into their corresponding
lemma form.

In addition, we analyzed manually the results obtained for a subset of
the extracted triples, looking at the sentences in the corpus where
they occurred. The subset contains 2,658 examples of four average
common verbs in the Treebank: {\it rise}, {\it report}, {\it seek} and
{\it present} (from now on, the {\it testing sample}).  On the one
hand, 235 (8.8\%) of these triples were considered to be extracted
erroneously because of the parser, and 51 (1.9\%) because of the
lemmatizer. Summarizing, 2,372 (89.2\%) of the triples in the testing
set were considered to be correctly extracted and lemmatized.

When analyzing the coverage of Word\-Net ta\-xo\-no\-my\footnote{The
\mbox{information of proper nouns} in WordNet is poor. For
\mbox{this reason we assign four predefined classes} to them:
\mbox{$<\!person, individual\!>$}, \mbox{$<\!organization\!>$},
\mbox{$<\!administrative\_district\!>$} and
\mbox{$<\!city\!>$}.} we considered two different ratios. On the one
hand, how many of the noun occurrences have one or more senses
included in the taxonomy: 113,583 of the 117,215 extracted triples
(96.9\%). On the other hand, how many of the noun occurrences in the
testing sample have the correct sense introduced in the taxonomy:
2,165 of the 2,372 well-extracted triples (92.3\%). These figures give
a positive evaluation of the coverage of WordNet.

In order to evaluate the performance of the learning process we
inspected manually the SRs acquired on the testing-sample, assessing
if they corresponded to the actual SRs imposed.  A first way of
evaluation is by means of measuring {\it precision} and {\it recall}
ratios in the testing sample. In our case, we define precision as the
proportion of triples appearing in syntactic positions with acquired
SRs, which effectively fulfill one of those SRs. Precision amounts to
79.2\%. The remaining 20.8\% triples didn't belong to any of the
classes induced for their syntactic positions. Some of them because
they didn't have the correct sense included in the WordNet taxonomy,
and others because the correct class had not been induced because
there wasn't enough evidence.  On the other hand, we define recall as
the proportion of triples which fulfill one of the SRs acquired for
their corresponding syntactic positions. Recall amounts to 75.7\%.

A second way of evaluating the performance of the abstraction process
is to manually diagnose the reasons that have made the system to
deduce the SRs obtained.  Table \ref{seek-1-table} shows the SRs
corresponding to the {\it subject} position of the verb {\it
seek}. {\it Type} indicates the diagnostic about the class
appropriateness. {\it Assoc}, the value of the association score. ``\#
$n$'', the number of nouns appearing in the corpus that are contained
in the class. Finally, ``\# $s$'' indicates the number of actual noun
senses used in the corpus which are contained in the class. In this
table we can see some examples of the five types of manual diagnostic:

\begin{description}

\item[Ok] The acquired SR is correct according to the noun senses
contained in the corpus.

\item[$\Uparrow$Abs]
The best level for stating the SR is not the one induced, but a lower
one.  It happens because erroneous senses, metonymies, ..., accumulate
evidence for the higher class.

\item[$\Downarrow$Abs]  Some of the SRs could be best gathered in a
unique class. We didn't find any such case. \\

\item[Senses] The class has cropped up because it accumulates enough
evidence, provided by erroneous senses.

\item[Noise] The class accumulates enough evidence provided by
erroneously extracted triples.

\end{description}

Table \ref{results-table} shows the incidence of the diagnostic types
in the testing sample.  Each row shows: the type of diagnostic, the
number and percentage of classes that accomplish it, and the number
and percentage of noun occurrences contained by these classes in the
testing sample
\footnote{this total doesn't equal the number of triples in the
testing sample because the same noun may belong to more than one class
in the final SRs}.  Analyzing the results obtained from the testing
sample (some of which are shown in tables \ref{seek-1-table} and
\ref{results-table}) we draw some positive ({\sl a}, {\sl e}) and some negative
conclusions ({\sl b}, {\sl c}, {\sl d} and {\sl f}):
\begin{description}
\item[{\sl a}.] Almost one correct semantic class for each syntactic position
in
the sample is acquired. The technique achieves a good coverage, even
with few co-occurrence triples.

\item[{\sl b}.] Although many of the classes acquired result
from the accumulation of incorrect senses (73.3\%), it seems that
their size tends to be smaller than classes in other categories, as
they only contain a 51.4\% of the senses .

\item[{\sl c}.] There doesn't seem to be a clear
co-relation between Assoc and the manual diagnostic. Specifically, the
classes considered to be correct sometimes aren't ranked in the higher
positions of the Assoc (e.g., Table \ref{seek-1-table}).

\item[{\sl d}.] The over-generalization seems to be produced because
of little difference in the nouns included in the rival classes.
Nevertheless this situation is rare.

\item[{\sl e}.] The impact of noise provided by  erroneous extraction of
co-occurrence triples, in the acquisition of wrong semantic classes,
seems to be very moderate.

\item[{\sl f}.]  Since different verb senses occur in the corpus,
the SRs acquired appear mixed.

\end{description}

\begin{table}
\centering

\begin{tabular}{||c|r|r|r|r||} \hline
{\it Diagnostic} & \# $Classes$ & {\it \%} & \# $n$ & {\it \%} \\
\hline Ok & 45 & 18.8 & 2,099 & 39.4 \\ $\Uparrow$Abs & 7 & 2.9 & 362
& 6.8 \\ $\Downarrow$Abs & 0 & 0.0 & 0 & 0.0 \\ Senses & 176 & 73.3 &
2,740 & 51.4 \\ Noise & 12 & 5.0 & 130 & 2.4 \\ \hline Total & 240 &
100.0 & 5,331 & 100.0 \\ \hline
\end{tabular}

\caption{Summary of the SRs acquired}
\label{results-table}

\end{table}

\section{\bf \sc Further Work}
\label{conclusions}

Although performance of the technique presented is pretty good, some
of the detected problems could possibly be solved. Specifically, there
are various ways to explore in order to reduce the problems stated in
points {\sl b} and {\sl c} above:

\begin{enumerate}

\item To measure the Assoc by means of Mutual Information between the
pair {\it v-s} and {\it c}. In this way, the syntactic position also
would provide information (statistical evidence) for measuring the
most appropriate classes.

\item To modify the Assoc in such a way that it was based in a
likelihood ratio test \cite{dunning-93}. It seems that this kind of
tests have a better performance than mutual information when the
counts are small, as it is the case.

\item To estimate the probabilities of classes, not directly from the
frequencies of their noun members, but correcting this evidence by the
number of senses of those nouns, e.g
\[ P(c|s) \approx
\frac{\sum_{n \in c} count(v,s,n) \frac{\#senses(n) \in c}{\#senses(n)}}
        {\sum_{v' \in {\cal V}} \sum_{n' \in {\cal N}} count(v',s,n')}
\]
In this way, the estimated function would be a probability
distribution, and more interesting, nouns would provide evidence on
the occurrence of their hyperonyms, inversely proportional to their
degree of ambiguity.

\item To collect a bigger number of examples for each verbal
complement, projecting the complements in the internal arguments,
using diathesis sub-categorization rules. Hopefully, Assoc would have
a better performance if it was estimated on a bigger population. On
the other hand, in this way it would be possible to detect the SRs
holding on internal arguments.

\end{enumerate}

In order to solve point {\sl d} above, we have foreseen two
possibilities:

\begin{enumerate}

\item To take into
consideration the statistical significance of the alternatives
involved, before doing a generalization step, climbing upwards,

\item To use the PPs that in the
corpus are attached to other complements and not to the main verb as a
source of ``implicit negative examples'', in such a way that they
would constrain the over-generalization.

\end{enumerate}

Finally, It would be interesting to investigate the solution to point
{\it f}.  One possible way would be to disambiguate the senses of the
verbs appearing in the corpus, using the SRs already acquired and
gathering evidence of the patterns corresponding to each sense by
means of a technique similar to that used by
\cite{yarowsky-92}. Therefore, once disambiguated the verb senses it
would be possible to split the set of SRs acquired.

Some of the future lines of research outlined above have been already
investigated and their results included in \cite{ribas-wp94}.

\newcommand{\etalchar}[1]{$^{#1}$}

\end{document}